\documentclass[letterpaper, 10 pt, conference]{ieeeconf}
\IEEEoverridecommandlockouts
\usepackage{cite}
\usepackage{array}
\newcolumntype{P}[1]{>{\centering\arraybackslash}p{#1}}
\usepackage{booktabs}
\usepackage{censor}
\usepackage{hyperref}
\usepackage[belowskip=-4pt,aboveskip=0pt]{caption}

\usepackage[pagewise]{lineno}
\usepackage{graphicx}
\graphicspath{{./figures/}}
\usepackage{subfig}

\title{\LARGE \bf
DeltaZ: An Accessible Compliant Delta Robot Manipulator for Research and Education
}

\author{Sarvesh Patil$^{*1}$, Samuel C. Alvares$^{*2}$, Pragna Mannam$^{1}$,  Oliver Kroemer$^{1}$,  F. Zeynep Temel$^{1}$
\thanks{$^{*}$Indicates equal contribution.}
\thanks{$^{1}$S. Patil, P. Mannam, O. Kroemer, and F. Z. Temel are with the Robotics Institute, Carnegie Mellon University, 
        Pittsburgh, PA 15213, USA
        {\tt\small \{sarveshp, pmannam, okroemer, ztemel\} @andrew.cmu.edu}}%
\thanks{$^{2}$S.C. Alvares is with the Robotics Institute Summer Scholars at Carnegie Mellon University, Pittsburgh, PA 15213, USA and also with Rose-Hulman Institute of Technology, Terre Haute, IN 47803, USA
        {\tt\small alvaresc@rose-hulman.edu}} \\
\authorblockA{\url{https://github.com/ZoomLabCMU/DeltaZ}}}

\begin{document}

\maketitle
\thispagestyle{empty}
\pagestyle{empty}

\begin{abstract}
This paper presents the DeltaZ robot, a centimeter-scale, low-cost, delta-style robot that allows for a broad range of capabilities and robust functionalities. Current technologies allow DeltaZ to be 3D-printed from soft and rigid materials so that it is easy to assemble and maintain, and lowers the barriers to utilize. Functionality of the robot stems from its three translational degrees of freedom and a closed form kinematic solution which makes manipulation problems more intuitive compared to other manipulators. Moreover, the low cost of the robot presents an opportunity to democratize manipulators for a research setting. We also describe how the robot can be used as a reinforcement learning benchmark. Open-source 3D-printable designs and code are available to the public.
\end{abstract}

\begin{keywords} 
Education Robotics, Parallel Robots, Compliant Joints and Mechanisms, Additive Manufacturing, Soft Robot Applications, Flexible Robotics, Kinematics
\end{keywords}

\section{Introduction}
As the manipulation capabilities of robots increase, new application domains will continue to emerge in semi-structured environments such as homes, warehouses, and hospitals. Given this potential impact on daily living, it is important that the field of robotic manipulation becomes accessible to a broader range of students and researchers. This need for accessibility is further strengthened by manipulation challenges becoming popular test beds for machine learning algorithms, which are already ubiquitous. A key barrier to exploring robotic manipulation is accessibility to hardware. Low-cost educational robots often focus on mobility rather than manipulation, and low-cost hardware for research is often still thousands of dollars and thus poses a significant barrier. 

\begin{figure}[t]
\setlength\belowcaptionskip{-25pt}
    \centering
    \includegraphics[width=\columnwidth,keepaspectratio]{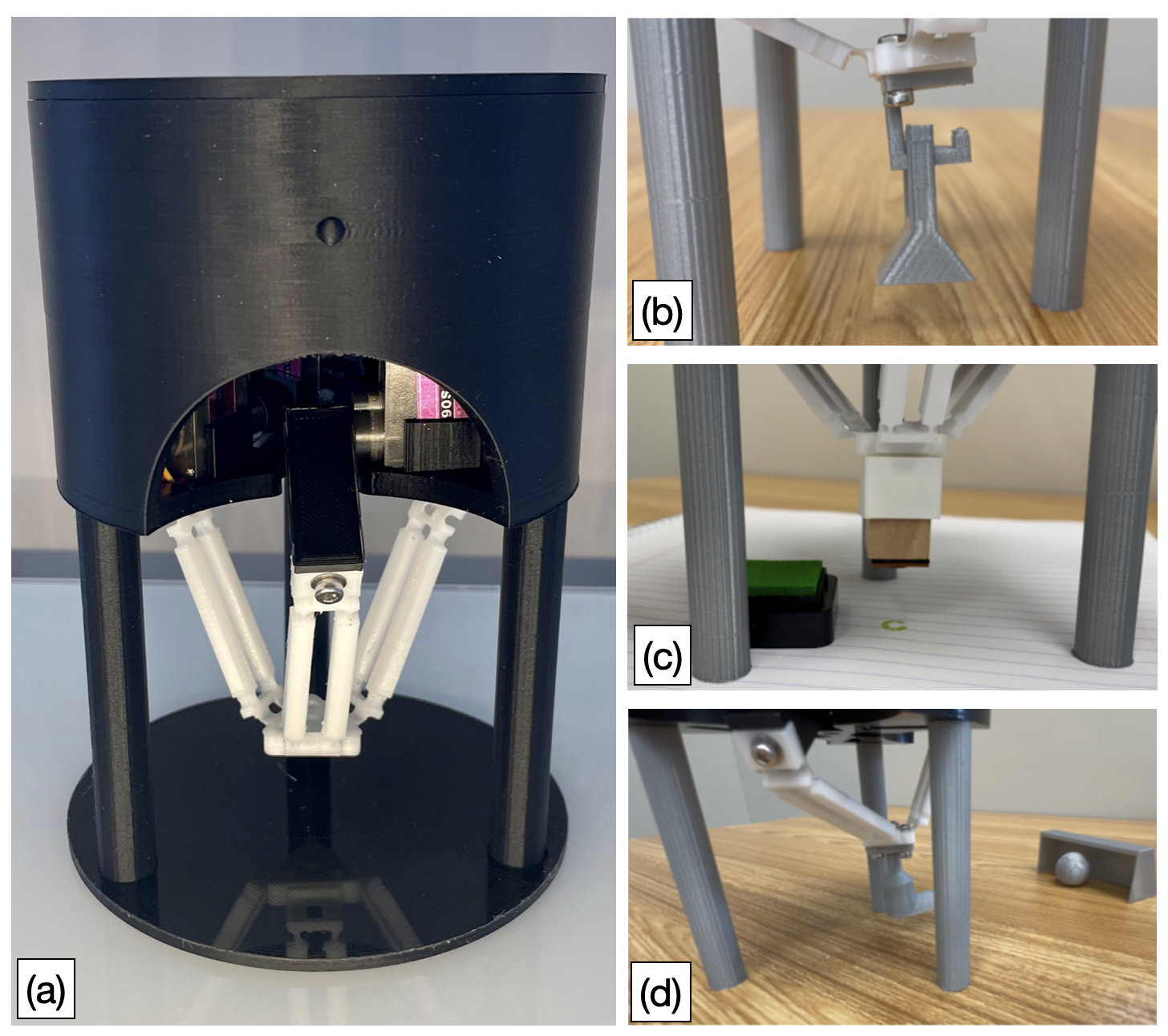}
    \caption{The DeltaZ and some of its functionalities. DeltaZ is a low-cost manipulation robot with 3-D printed mechanical parts and a compliant end-effector (a). DeltaZ's hook end-effector picking up a weight (b), with a stamp tool and an ink pad, stamping a letter (c), and  with a 3D-printed boot attachment, kicks a soccer ball into the goal (d).}
    \label{fig:figure1}
\end{figure}

To democratize robotic manipulation for hands-on education and research, we propose a low-cost, open-source, 3D-printed delta robot design, DeltaZ,  shown in Fig. \ref{fig:figure1}. This robot costs around $50$ USD and its  design allows for it to be easily assembled and controlled. A core part of the design is the three 4-bar mechanism links and articulated platform that are 3D printed as a single piece using a soft material such as thermoplastic polyurethane (TPU). This structure allows the robot to safely interact with its surroundings while still achieving high repeatability. The delta kinematics also allow the robot to directly translate the end-effector in 3D Cartesian space, with only small rotations, which allows the three degrees-of-freedom (DoF) robot to be used for a wide range of manipulation tasks. Being able to easily control the 3D position also provides a more intuitive experience for novice students when compared to three rotational DoF serial kinematics. The compliant delta structure also allows the robot to be used for dynamic tasks, such as striking a ball or puck. The base plate can be easily replaced with different designs to create various task-specific environments and structures, and additional sensors can  be easily incorporated to augment the robot's capabilities. 

In the following sections, we present the design of the low-cost delta robot and its fabrication. We evaluate the robot's capabilities in terms of speed and repeatability. Our experiments demonstrate the robustness of the robot and show how it can be used as a benchmarking tool for robot reinforcement learning, as well as a basis for multi-robot transfer learning. 

\section{Related Work}

\subsection{Delta Robots}
Delta robots are parallel robots with 3 translational degrees of freedom where the end-effector stays parallel to its base (as shown in Fig.\ref{fig:figure3}(a)). The motors are stationed in the body which allows the light end-effectors to perform pick and place tasks with high accuracy and precision  \cite{merlet2005parallel}. With closed-form inverse kinematic solutions, the precision of delta robots can be controlled sufficiently for biomedical and surgery applications \cite{mcclintock_millidelta_2018}. Previous work shows 3D-printed compliant end-effectors with linearly actuated delta robots can be used for dexterous manipulation \cite{RSSdelta}. We build upon this work to make a compliant delta robot manipulator with revolute actuation for safe interactions with objects and people. 

\subsection{Educational Robots}
A number of educational robotics kits have been developed in recent years \cite{sapounidis_educational_2020}. These robots are often designed to teach students about coding and make the sense-plan-act loop accessible and engaging to K-12 students. However, these robots tend to emphasize mobility rather than manipulation, with manipulation often being a 1 DoF extension to the robot like a fork lift. The focus on mobility also means that the lesson is often on the robots avoiding contacts, or only allowing for a narrow set of interactions, rather than using compliance to exploit contacts with the environment. Our DeltaZ robot was designed to afford safe compliant interactions. Future goals include developing the teaching material that one would need for a full educational kit and course. 

\subsection{Benchmarking for Robot Manipulation}

Developing benchmarking tasks for robot manipulation research is challenging, and a number of workshops at international conferences have been dedicated to this issue \cite{calli2015benchmarking, ieeeRAS, iros2021WS}. A core problem is the need to have similar objects and hardware to compare different algorithms without having to repeat entire experiments. Object sets and shared experimental protocols have helped to make experiments more reproducible \cite{YCB}. However, differences in robot hardware and low-level control can still have a significant effect on results. 

To avoid hardware issues, simulators have become ubiquitous tools for evaluating robot manipulation algorithms \cite{metaworld}. However, modern simulation engines still exhibit a large simulation-to-reality gap and often provide an over-idealized version of tasks. The simulation-to-reality gap is exacerbated when modeling the complexities of manipulating non-rigid objects. The DeltaZ robot makes real-robot evaluations more accessible, allowing for easier hands-on experimentation.

Remote experimentation sites, wherein the robots are controlled remotely over the internet, provide access to robots for a larger population of researchers \cite{gym_duckietown}. These sites allow state of the art robots to be accessed from around the globe. However, the remote nature of these tasks makes it difficult to create new task environments. For the DeltaZ robot, new task environments and objects can be 3D printed given its scale. The DeltaZ also encourages hands-on research which can often lead to new insights when observing experiments in person. 

It should also be noted that simulations and remote experimentation sites are favored by some researchers, and especially non-robotics researchers, as they avoid the need for setting up and maintaining a physical robot system. These processes can be time consuming and often require additional engineering expertise that may not be present in some labs. The DeltaZ robot provides a simple tabletop environment that is largely self contained and easy to assemble and maintain.

\subsection{Low-cost Research Manipulators}
The need for accessible research robots has resulted in a number of low-cost robots being developed in recent years. For example, the Locobot incorporates a 5 DoF robot arm, and the Dynamixel Claw includes finger-like 3 DoF manipulators\cite{murali2019pyrobot, BAIRblog}. These manipulators use serial kinematic designs, like traditional robot arms. Serial kinematics result in each motor affecting the end-effector's position and orientation. A robot will therefore often need to have additional degrees of freedom simply to maintain a certain orientation, or the task will need to be designed to reduce the effects of the rotations (e.g., using a spherical end-effector). The delta design allows us to create and perform a wide range of translational manipulation tasks with a simple 3 DoF design.

A serial design also means that the servos need to be strong enough to support the other motors in the chain. This requirement does not only increase the cost of the motors, but can also be taxing on the motors over time. The parallel design of the delta robots allows us to use lower-cost servo motors. 

The cost of typical robotic manipulators is in the range of thousands of dollars \cite{BAIRdynamixel, onrobot, robotiq}. 
To make the DeltaZ accessible to students and researchers, we designed the 3D-printed revolute delta robot to cost around $50$ USD.

\section{DeltaZ Robot Design}

The core design goals of the DeltaZ are to make a robot that is precise, versatile, low-cost, and can withstand impacts and obstacles due to its compliance. The affordable robot enables a multitude of manipulation tasks to be achieved and is ideal for an educational or research setting. 
Mechanical design, electronics and control, as well as examples of robot functionalities are discussed in this section. 

\subsection{3D-Printing and Component Overview}
\begin{figure}
    \centering
    \includegraphics[width=\columnwidth]{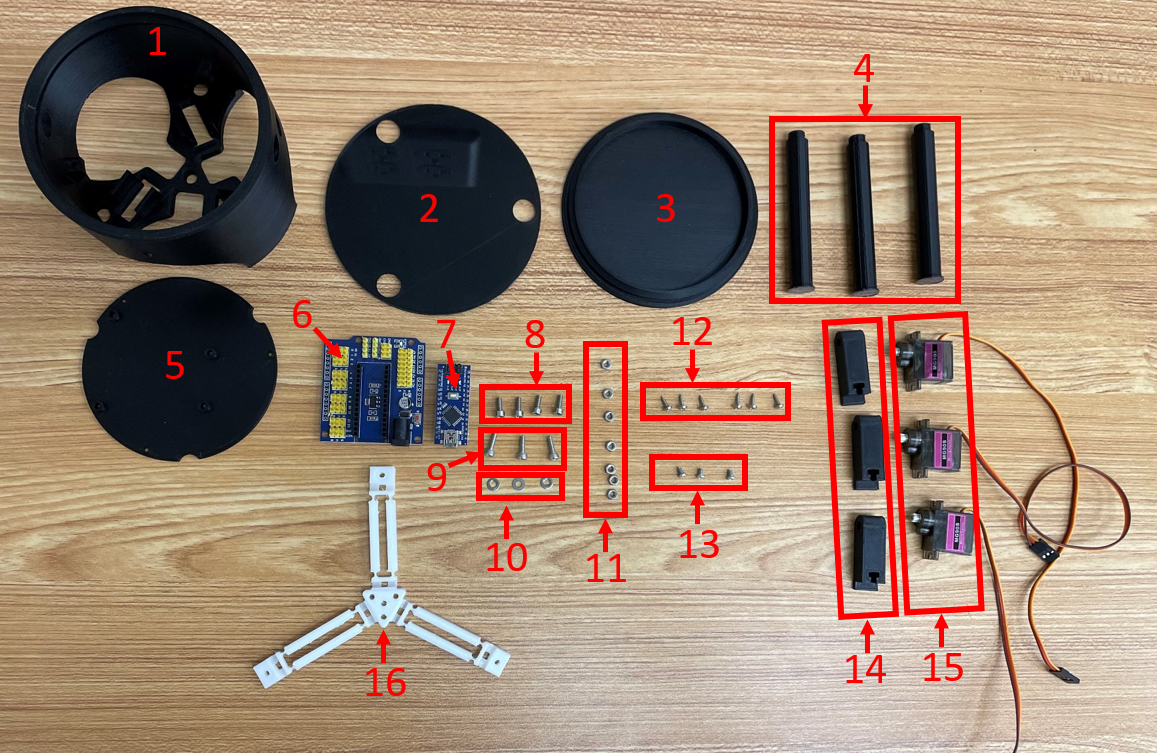}
    \caption{All of the parts required to build the robot, including the 3D-printed parts. Parts include (1) Body, (2) Base, (3) Cap, (4) Leg, (5) Divider, (6) Expansion Board, (7) Nano, (8) 8 mm Screw, (9) 10 mm Screw, (11) Nut, (12) Self Tapping Screw, (13) Servo Screw, (14) Forearm, (15) Servo, (16) Compliant End-Effector. Note that Self Tapping Screws (12) and the Servo Screws (13) are included with the Servos (15).}
    \label{fig:numParts}
\end{figure}

\begin{figure}
    \centering
    \includegraphics[width=\columnwidth]{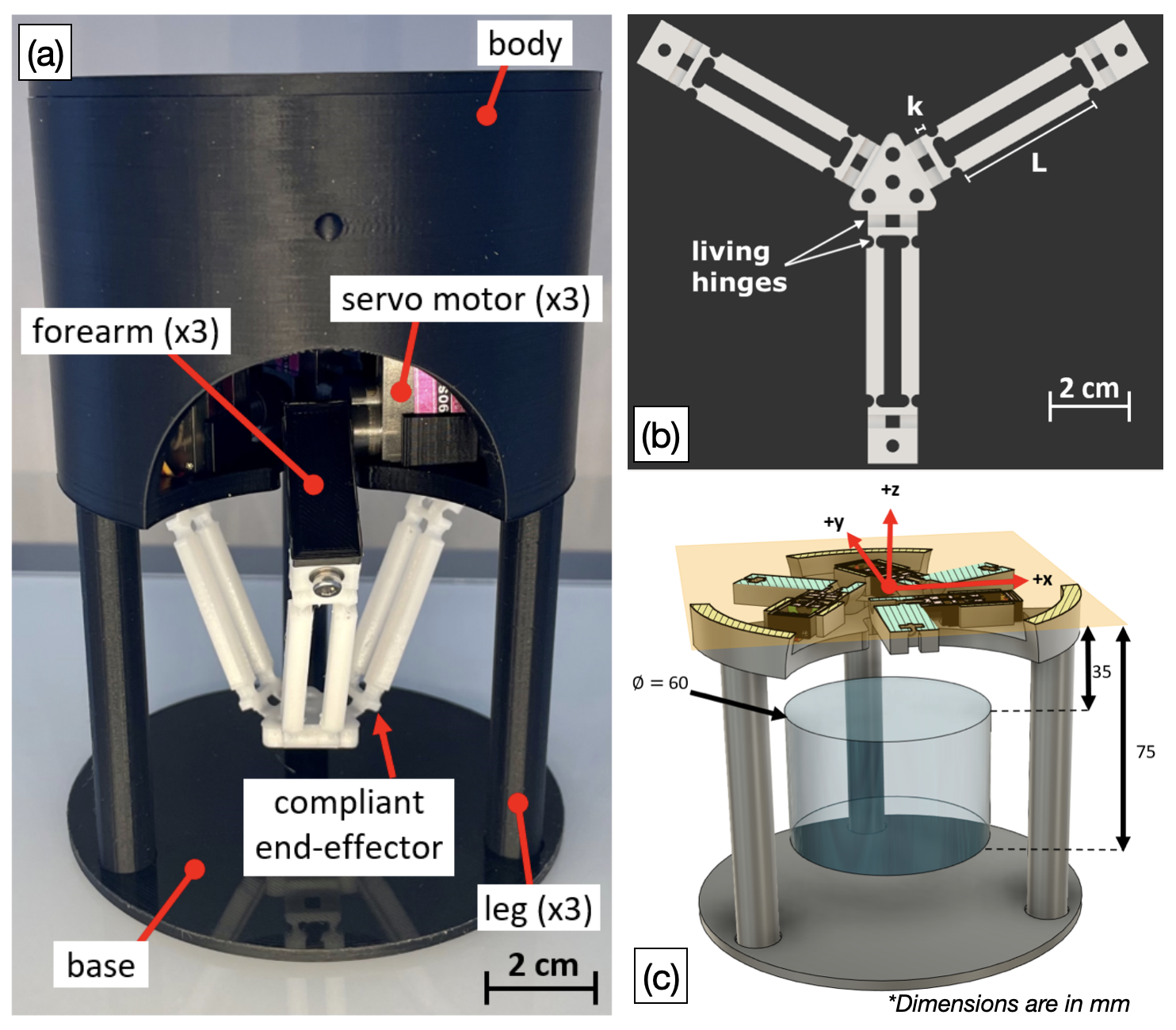}
    \caption{(a) Assembled DeltaZ. The major parts include the body, servo motor, forearm, 
    compliant end-effector (printed with white material), legs, and base. The body encloses all motors and electronics and is supported by three legs which form a tripod. (b) 
    The compliant end-effector of DeltaZ can be 3D-printed with two orthogonal revolute joints printed as living hinges. Parameters like k and L can be modified by users. The four central holes to attach a variety of end-effectors using M3 screws. This component bolts to the rigid forearms via the outer three screw holes. (c) DeltaZ's allowable workspace shown as a transparent cylindrical solid with a diameter of 60mm and height of 75 mm. }
    \label{fig:figure3}
\end{figure}

DeltaZ is made of a total of 
42 pieces, including individual screws. An overview of the individual pieces is shown in Fig.\ref{fig:numParts}, and assembled and labeled in  Fig.\ref{fig:figure3}. The black pieces (components 1-5 and 14 in Fig.\ref{fig:numParts}) are all 3D printed from Polylactic Acid (PLA). These components include a housing to encase the electronics and mount the servos, as well as a set of legs and a base plate for supporting the delta robot's body when used in a top-down manner. 

The mechanical assembly using screws and bolts provides modularity, i.e., all of the components can be replaced if needed as nothing is glued. The time to assemble the robot varies based upon the users experience level, but can generally be done within an hour. A video tutorial on DeltaZ's website \cite{sam_alvares_building_2021} walks users through the assembly process. 

All designs are open hardware which allows communities of users to share ideas and collaborate on adapting the overall design and functionality of the robot to specific applications \cite{noauthor_open_nodate}.

\subsection{Soft Mechanisms and Living Hinges}

The white component (16) in Figs. \ref{fig:numParts}, \ref{fig:figure3}(a), and \ref{fig:figure3}(b) is referred to as the compliant end-effector, as it is printed from a flexible material, such as TPU 95A. Delta robots use a set of parallel 4-bar mechanisms (parallelograms) to maintain the end-effector's orientation. This part of the design often results in a large number of additional components that increase the overall complexity of the robot's design. To provide a design that is simple, yet precise, we 3D print the entire structure as a single print. 

The flexible joints are made as living hinges to mimic a single revolute joint. A combination of two living hinges performs similarly to a universal joint, which is typically present in a conventional delta robots. Living hinges are made by locally reducing material thickness to a compliant articulated joint.
For experiments, we printed delta parts using Ultimaker TPU 95A, PP (polypropelene), and PLA (polylactic acid) material. From most to least rigid, the tensile moduli are 2,346.5 MPa (using ISO 527), 220 MPa (using ISO 527), 26 MPa (using ASTM D638) for PLA, PP, and TPU, respectively \cite{datasheets}. The materials were chosen for their ability to be 3D-printed and create living hinges at the desired locations. Non-rigid materials like PP and TPU, in addition to relatively low-torque motors, make the robot safe to users. 

Design parameters like parallelogram beam and hinge thickness were chosen based on prior work \cite{siciliano_characterization_2021}. The beam thickness is $4.5$ mm for TPU and PLA deltas and $2.5$ mm for PP delta parts as in \cite{RSSdelta, siciliano_characterization_2021}. The hinges are all $0.41$ mm thick which was as close to the desired $0.375$ mm found in previous work but could also be printed on various 3D-printers (limited by size of material extruded).
Parameters L and k in Fig. Fig.\ref{fig:figure3}(b) are parallelogram link lengths of $L=37$ mm, and $k=5.25$ mm which is the offset between the two orthogonal revolute joints made from living hinges. These two revolute joints should be as close as possible ($k\rightarrow0$ mm) to minimize positional error \cite{RSSdelta}. Using key physical parameters from previous work, we improved upon the mechanical design of a compliant, delta-style robot by making it more conducive for a learning and research environment. In particular, mechanical connections were made more durable and reliable. Also, simple assembly and fabrication processes were formalized and made open-source. Users may experiment with the hinge thickness, parallelogram link length, and offset between revolute joints to test the affect on kinematic behavior of the compliant end-effector. Additionally, changes in the resulting workspace could be helpful for specific manipulation tasks.



\subsection{Base Plate}
The base  plate (part 2 in Fig. \ref{fig:numParts}) serves as a support surface for the robot's manipulation tasks. The basic plate is flat and simply ensures that the robot cannot push off of the underlying table. However, the plate can also be easily replaced with other base plates to create different task-specific environments. For our experiments, one of the base plates includes a potentiometer mounting for a dial turning experiment. Other designs could including fixtures for different tasks. In this manner the robot can be easily adapted to explore different tasks and provide a self-contained environment for benchmarking tasks. Switching out the base plate can be done in a matter of minutes by simply unscrewing the legs, exchanging the plates, and reattaching the legs. The base plate can also be removed to allow the robot to perform simple locomotion tasks with a single articulated leg.

\subsection{Arduino, Servos, and Sensorization}
DeltaZ is driven by an Arduino Nano microcontroller and powered via USB. Delta robots have base mounted motors and parallel geometry that allow for fast and accurate motions with relatively small and low-cost motors \cite{lopez_delta_2006, mcclintock_millidelta_2018}. Thus, we are able to articulate DeltaZ using affordable, 9-gram, metal geared, micro servo motors. DeltaZ can be positioned by both forward and inverse kinematics in open-loop control. The Arduino also allows for simple sensors, such as buttons, light sensors, or potentiometers, to be easily incorporated into the platform. 


\subsection{Serial Interface}

The robot is controlled externally through the serial port interface of the Arduino. The interface allows the desired angles to be directly specified or the desired x-y-z location of the end-effector to be given. For the latter, the inverse kinematics are computed directly on the Arduino to compute the corresponding desired angles based on a rigid-link model of the robot. We limit the workspace to a cylindrical region of height $40$ mm and diameter 60 mm, as shown in Fig. \ref{fig:figure3}(c), such that the end-effector cannot collide with the legs of the robot. The interface can also be used to read the values of sensors connected to the Arduino. All software and hardware designs are open-source.

\section{DeltaZ for RL Benchmarking}
To show the efficacy of the DeltaZ for real-robot benchmarking, we have the robot apply reinforcement learning to acquire a dial turning skill. The learning processes is repeated multiple times across three different copies of the robot to demonstrate the similar outcomes.

\subsection{Dial Turning Task}

\begin{figure}[t]
\centering
\includegraphics[width=0.9\columnwidth,keepaspectratio]{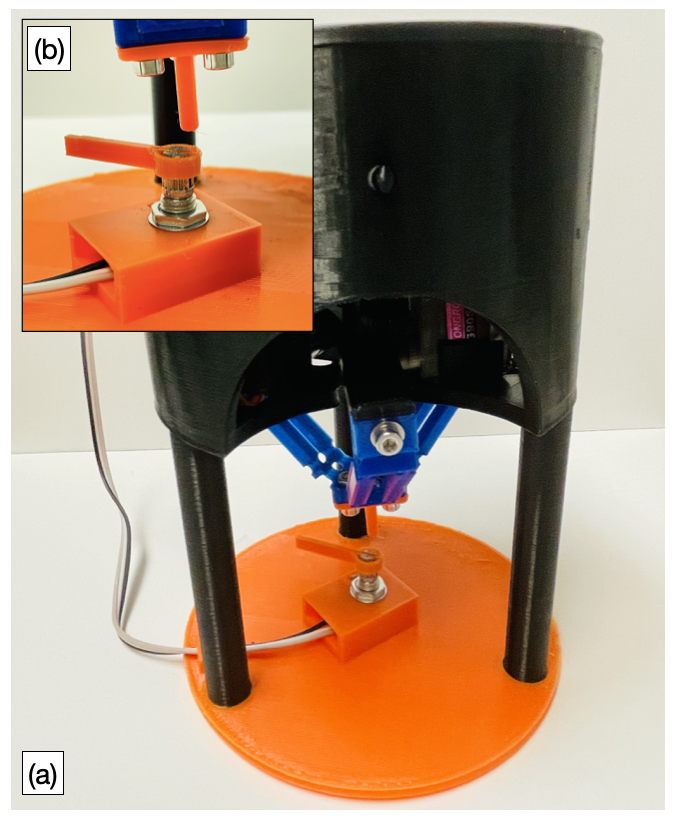}
\caption{DeltaZ robot with a potentiometer mounted on the base (a). The potentiometer (b) is connected to an analog input pin on the Arduino. Corresponding to the resistance value of the potentiometer, we obtain values from 0 to 1023 as analog input from the Arduino.}
\label{fig:deltaZ_pot}
\end{figure}

The example benchmarking task is designed around a potentiometer mounted in the base of the robot, as shown in Fig. \ref{fig:deltaZ_pot}. The goal of the task is for the robot to use its end-effector to turn the potentiometer to match a desired resistance value. A small lever has been attached to the potentiometer for the robot to push against. The potentiometer is mounted in a 3D-printed base plate that was designed for this task, and its pins are connected to the Arduino such that its resistance values can be easily measured by the robot. This task requires the robot to operate an articulated object through contact-based interactions, with different amounts of force required depending on where on the lever the robot pushes. The task was inspired by similar tasks used for the Dynamixel claw \cite{dynamixal}. For each epsiode, the robot receives a reward of $R=100$ if the final angle $\phi$ is within $15^{\circ}$ of the desired angle $\phi_d$, indicating a successful task completion, as well as a  quadratic cost based on the difference between the final angle and the desired angle $R=100[\|\phi-\phi_d\|<15]-10^{-5} (\phi-\phi_d)^{2}$.

To minimize the amount of human effort in running the evaluations, we incorporate an automatic resetting procedure for re-positioning the dial between trials. The end-effector goes behind the lever and rotates it back to the starting angle, at  which point a new episode can be executed.

The experiments were conducted across three different robots, with seven full learning processes run on each robot. The resulting 21 trails were performed to  demonstrate the reproducibility of the learning process across different robots.

\subsection{Skill Parameterization}
The end-effector of the DeltaZ robot is restricted to  a $30$ mm radius around the origin. Thus, in order to efficiently represent the action space, we parameterize it in the polar $(\rho, \theta)$ domain, where $\rho$ is the radius and $\theta$ is the angle from the positive x-axis. The z position of the end-effector is fixed during the skill execution at a height that it can push the potentiometer's lever.

Each skill execution is then defined by two points $(\rho_1, \theta_1)$, and  $(\rho_2, \theta_2)$ that define two waypoints. The robot moves down at the first waypoint and then moves across to the second waypoint in a straight line. Thus, the waypoints are defined in polar coordinates, but the trajectory itself is in a straight Cartesian x-y line.

The goal of the reinforcement learning is to learn a set of skill parameters $(\rho_1, \theta_1,\rho_2, \theta_2)$ for achieving a desired potentiometer angle.

\subsection{Skill Learning}
To demonstrate skill learning on the DeltaZ platform, we learn the dial turning task using episodic Relative Entropy Policy Search (eREPS) \cite{REPS}. eREPS models the distribution over the skill parameters as a 4D Gaussian distribution. We normalize the parameter values to be within a range of -1 to 1, and initialize the Gaussian distribution with mean $0.4$ and a diagonal covariance matrix with non-zero elements of $0.15$. 

As a model-free policy search approach, eREPS iterates between evaluating batches of sampled parameters on the real robot and updating the Gaussian policy based on the resulting rewards.  For our experiments, the robot rolls out 20 episodes initially, and then 10 episodes for subsequent iterations between each policy update. The actions sampled from the Gaussian distribution are passed to the Arduino which moves the end-effector to the desired location, reads the resistance value of the potentiometer, and sends it back to the computer to compute the reward. Given the automatic resets and potentiometer recording, the data collection process can be run fully autonomously. 

The policy updates of eREPS attempt to maximize the expected return while limiting the KL divergence between the old policy and the new policy. For our updates, the robot utilizes the most recent 20 samples for each policy update, i.e., from the last two policies. Retaining samples from multiple previous policies is a common practice for eREPS to further improve learning stability. The eREPS algorithm is terminated when all 10 trials in the previous batch were successful at completing the task.


\begin{figure}[h!]
\centering
\includegraphics[width=\columnwidth]{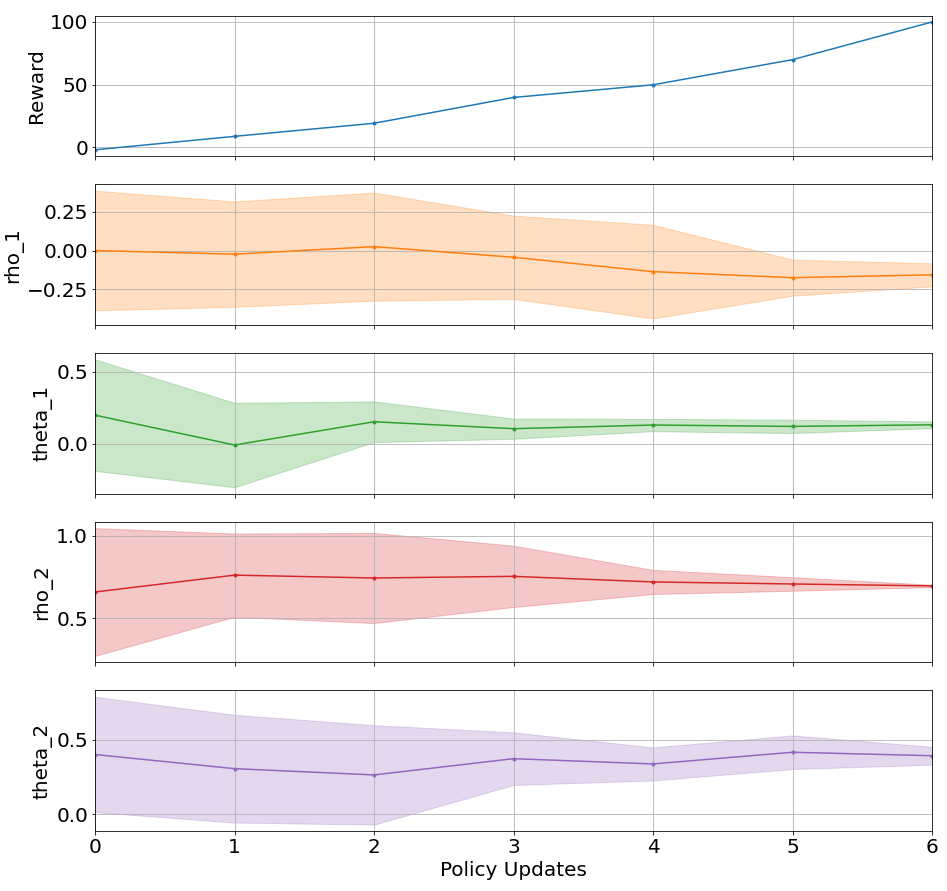}
\caption{Convergence of parameters for a single Robot.}
\label{fig:convergence}
\end{figure}


\begin{figure}
\centering
\includegraphics[width=\columnwidth,keepaspectratio]{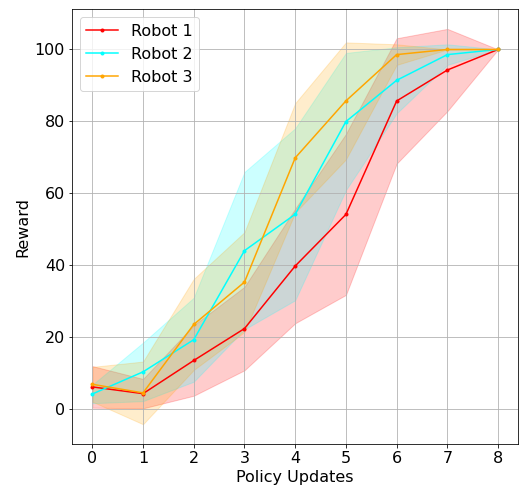}
\caption{Gaussian plots of rewards across different robots.}
\label{fig:rewards}
\end{figure}

\section{Evaluations}
Our experiments evaluated the performance of the robot performing tablet interaction tasks and the dial-turning RL benchmarking task. The results of our evaluations are given in this section.

\subsection{Drawing Task Evaluations}

\begin{figure}[t!]
\centering
\includegraphics[width=\columnwidth,keepaspectratio]{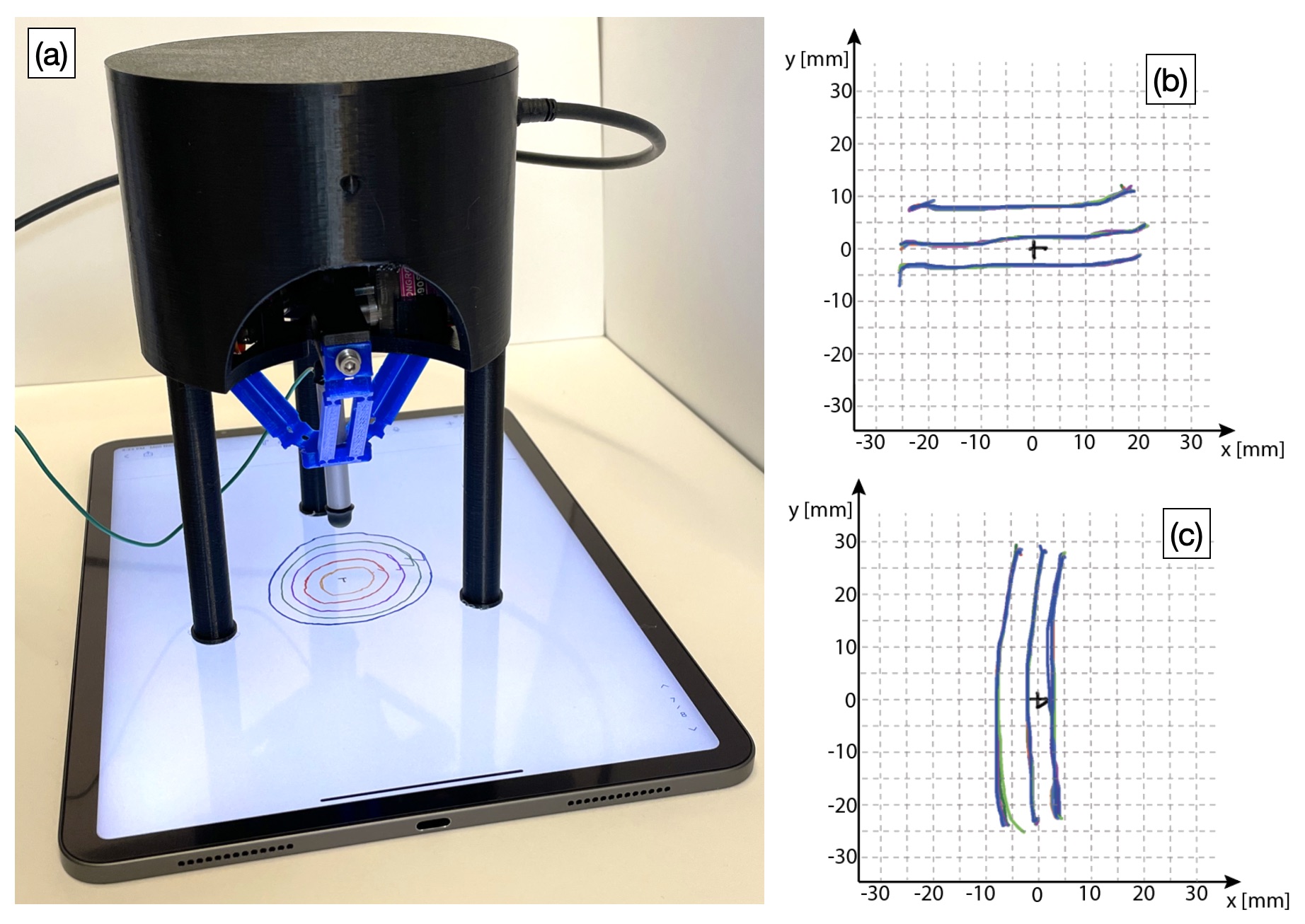}
\caption{DeltaZ is placed on an tablet with a stylus pen to characterize the workspace of the robot (a). The accuracy and precision of DeltaZ over 10 trials for straight lines along x (b) and y (c) axes are shown. 
}

\label{fig:delta_iPad}
\end{figure}

We used the DeltaZ robot to perform a series of drawing tasks to characterize its workspace. Two end-effectors of TPU and PP are used for the experiments. The end-effectors were equipped with a capacitive stylus pen and the drawing was performed on a tablet, as shown in Fig. \ref{fig:delta_iPad} (a). To test the repeatability of the DeltaZ, lines along x and y axes were drawn at different lateral and vertical distances. First, the robot is commanded to make a plus sign at the origin of the plane. Then, the stylus pen followed a straight trajectory along x-axis with three different y-values (Fig.~\ref{fig:delta_iPad} (b)). The same experiments are repeated to observe the accuracy and precision along the y-axis (Fig.~\ref{fig:delta_iPad} (c)). The results over 10 repetitions show that the accuracy of the robot is not high. However, it performs the tasks with very high precision. 

\begin{figure}[t!]
\centering
\includegraphics[width=\columnwidth,keepaspectratio]{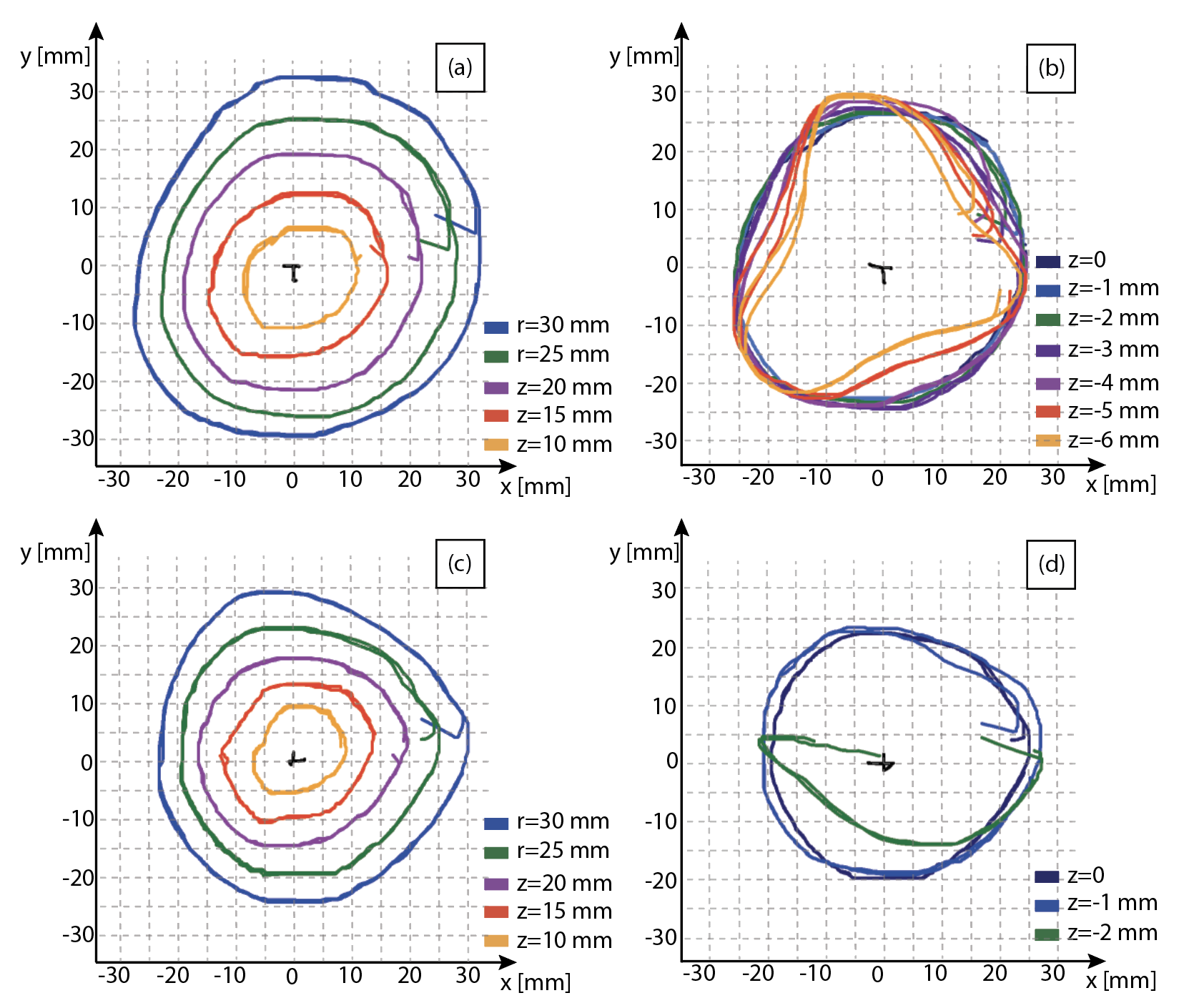}
\caption{The comparison between end-effectors that are printed with TPU (a and b) and PP (c and d). Radius of the circle varies between 30 mm and 10 mm, colors indicating different radii (a and c). As the distance between the stylus pen and the touchpad reduces, the shape changes due to the compliance of the end-effector (b and d). The effect of the material is clearly observed. Since the PP end-effector cannot conform to the environment as well as TPU end-effector, it snapped off of the forearms, hence resulting with a non-uniform shape (green line).}

\label{fig:TPUvsPP}
\end{figure}

The comparison between end-effectors printed using different materials is shown in Fig. \ref{fig:TPUvsPP} following a circular trajectory with different radii and different vertical distances. Both versions of DeltaZ were able to follow a fairly precise trajectory with a changing radius. However, as we increased the vertical distance and pushed the stylus pen into the tablet, we observed that the trajectory got distorted as the vertical distance of the TPU end-effector decreased. However, DeltaZ with the PP end-effector was not able to finish the tasks at different values along z-axis, due to the lack of compliance.


In order to measure the maximum speed of DeltaZ, we performed experiments where the end-effector follows a straight line of 40mm along the x-axis multiple times. The maximum speed at which DeltaZ was still able to follow the full trajectory is calculated as 0.17 m/s. 

\subsection{RL Benchmarking}
The eREPS algorithm was run on three different robots seven times each for learning the dial turning task. The goal of this experiment is to evaluate the reproducability of the results across the different robots in order to demonstrate their utility for benchmarking. 

The results of the experiments are shown in Fig. \ref{fig:convergence} and \ref{fig:rewards}. Fig. \ref{fig:convergence} shows an example of the average rewards and the distribution over the skill parameters for each policy update of one of the 21 learning processes. Fig. \ref{fig:rewards} shows the distribution over the seven average rewards for each of the robots. The error bars correspond to two standard errors. 

As shown in Fig. \ref{fig:convergence}, the robot starts with a broad Gaussian distribution to explore the parameter space. This initial exploration resulted in the end-effector colliding against the potentiometer axle several times. Due to the compliance of the end-effector, these collisions did not cause any damage and the robot could continue to perform the entire training process without human intervention.

The main result of this experiment is the similarity of the learning curves 
across the different robots. The means are close together, with a large amount of overlap between the standard error regions. This indicates that the evaluations performed on the different robots are comparable. These results demonstrate that the three different robots could be used for benchmarking algorithms. The ability to reproduce the task environment and run the experiment autonomously also allows for easier reproduction of results across robots.

As an additional evaluation, the learned skills were executed on the other robots, and we found that each of them succeeded using direct zero-shot transfer. This result demonstrates the potential of using the DeltaZ robots for multi-robot training and robot-robot transfer research in the future.

\vspace{-0.3cm}
\section{Conclusion}
We proposed the DeltaZ  design for creating an accessible robot platform for research and education. The 3D printed robot is easy to assemble and maintain, and it can be used to perform a variety of translational manipulation tasks. Our experiments demonstrated the robot's ability to perform precise repetitive tasks that exploit the robot's inherent compliance. We also demonstrated the utility of the robot as a benchmarking platform. In the future, we will explore extending the platform with additional sensors to perform various tasks, as well as create accompanying teaching materials to develop an educational kit.

\section{Acknowledgements}
The research presented in this work was funded by the
National Science Foundation under project Grant No. CMMI-2024794.

\nocite{*}  
\bibliographystyle{IEEEtran}
\bibliography{./IEEEfull,refs}

\end{document}